\documentclass{article}

    \PassOptionsToPackage{numbers, sort}{natbib}

    \usepackage[final]{neurips_2020}

\usepackage[utf8]{inputenc} 
\usepackage[T1]{fontenc}    
\usepackage{hyperref}       
\usepackage{url}            
\usepackage{booktabs}       
\usepackage{amsfonts}       
\usepackage{nicefrac}       
\usepackage{microtype}      

\usepackage{amsmath,amsfonts,bm}

\def\eqref#1{equation~\ref{#1}}

\def\1{\bm{1}}

\DeclareMathAlphabet{\mathsfit}{\encodingdefault}{\sfdefault}{m}{sl}
\SetMathAlphabet{\mathsfit}{bold}{\encodingdefault}{\sfdefault}{bx}{n}

\newcommand{\E}{\mathbb{E}}

\newcommand{\R}{\mathbb{R}}

\newcommand{\KL}{D_{\mathrm{KL}}}
\newcommand{\Var}{\mathrm{Var}}

\DeclareMathOperator*{\argmax}{arg\,max}
\DeclareMathOperator*{\argmin}{arg\,min}

\usepackage{booktabs}
\usepackage{enumitem}
\usepackage{mathtools}
\usepackage{amssymb}
\usepackage{bbm}
\usepackage{bm}
\usepackage{array}
\usepackage{multirow}
\usepackage{multicol}
\usepackage{makecell}
\usepackage{csquotes}

\usepackage{amssymb}
\usepackage{pifont}
\newcommand{\cmark}{\checkmark}

\usepackage[dvipsnames]{xcolor}

\usepackage{amsthm}

\newtheorem{definition}{Definition}

\title{Feature Removal Is A Unifying Principle For \\ Model Explanation Methods}

\author{
  Ian C. Covert \\
  University of Washington\\
  Seattle, WA \\
  \texttt{icovert@uw.edu} \\
   \And
  Scott Lundberg \\
  Microsoft Research\\
  Redmond, WA \\
  \texttt{scott.lundberg@microsoft.com} \\
   \And
  Su-In Lee \\
  University of Washington\\
  Seattle, WA \\
  \texttt{suinlee@uw.edu}
}

\begin{document}

\maketitle

\begin{abstract}
Researchers have proposed a wide variety of model explanation approaches, but it remains unclear how most methods are related or when one method is preferable to another. We examine the literature and find that many methods are based on a shared principle of \textit{explaining by removing}---essentially, measuring the impact of removing sets of features from a model. These methods vary in several respects, so we develop a framework for \textit{removal-based explanations} that characterizes each method along three dimensions: 1)~how the method removes features, 2)~what model behavior the method explains, and 3)~how the method summarizes each feature's influence. Our framework unifies 26 existing methods, including several of the most widely used approaches (SHAP, LIME, Meaningful Perturbations, permutation tests). Exposing the fundamental similarities between these methods empowers users to reason about which tools to use, and suggests promising directions for ongoing model explainability research.\footnote{Since its
initial publication, an extended version
of this work
was published in the Journal of Machine Learning Research~\cite{covert2021explaining}.}
\end{abstract}

\section{Introduction}

The proliferation of black-box models has made machine learning (ML) explainability an increasingly important subject, and researchers have now proposed a wide variety of model explanation approaches \cite{breiman2001random, chen2018learning, covert2020understanding, lundberg2017unified, owen2014sobol, petsiuk2018rise, ribeiro2016should, sundararajan2017axiomatic, vstrumbelj2009explaining, zeiler2014visualizing}. Despite progress in the field, the relationships and trade-offs among these methods have not been rigorously investigated, and researchers have not always formalized their fundamental ideas about how to interpret models~\citep{lipton2018mythos}. This makes the literature difficult to navigate and raises questions about whether existing methods relate to human processes for explaining complex decisions~\citep{miller2017explainable, miller2019explanation}.

\begin{figure*}[t]
\vskip 0.1in
\begin{center}
\includegraphics[width=\columnwidth]{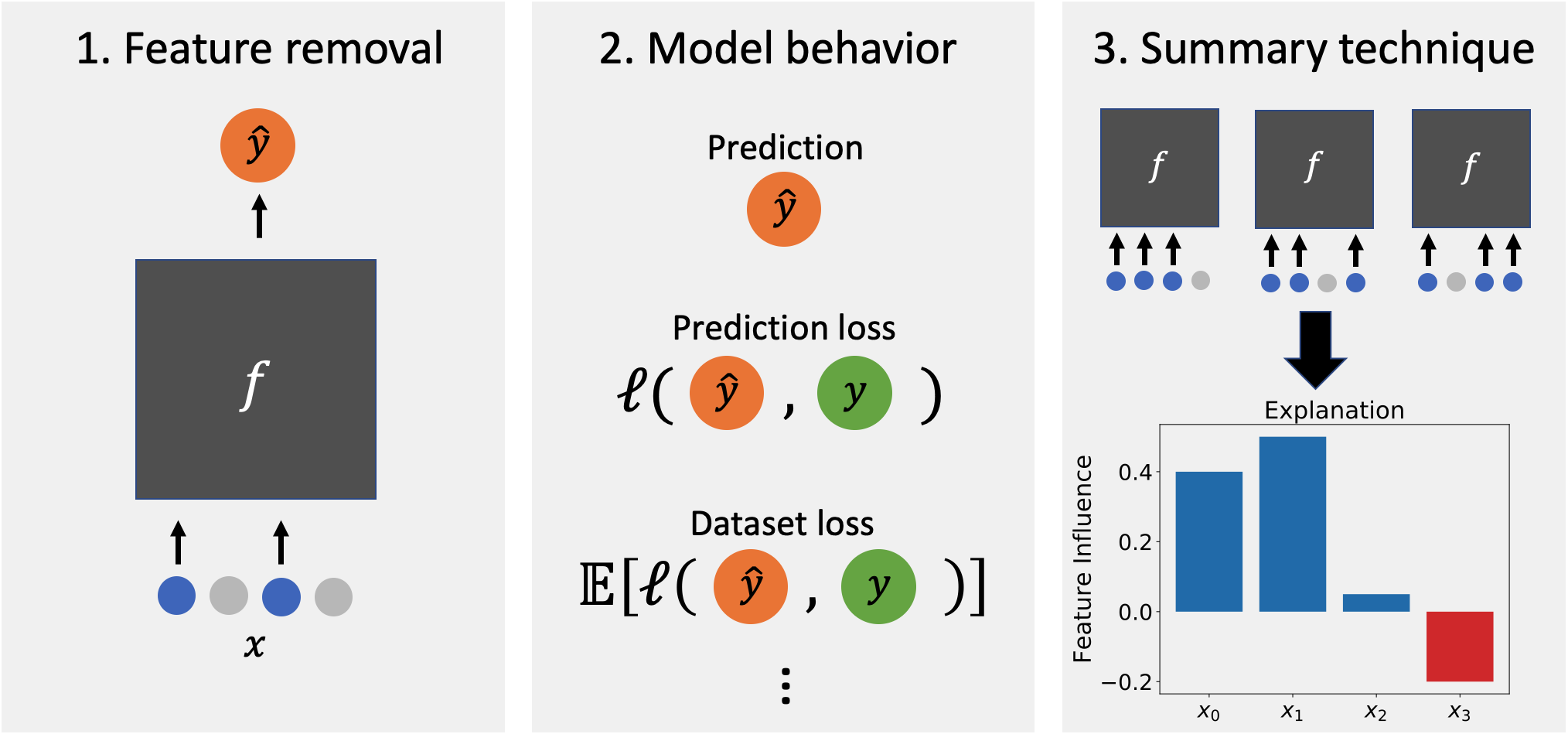}
\vskip 0.1in
\caption{A unified framework for \textit{removal-based explanations}. Each method is determined by three choices: how it removes features, what model behavior it analyzes, and how it summarizes feature influence.}
\label{fig:concept}
\end{center}
\vskip -0.1in
\end{figure*}

Here, we present a comprehensive framework that unifies a substantial portion of the model explanation literature. Our framework is based on the observation that many methods can be understood as \textit{simulating feature removal} to quantify each feature's influence on a model. The intuition behind these methods is similar (depicted in Figure~\ref{fig:concept}), but each one takes a slightly different approach to the removal operation: some replace features with neutral values~\citep{petsiuk2018rise, zeiler2014visualizing}, others marginalize over a distribution of values~\citep{lundberg2017unified, strobl2008conditional}, and still others train separate models for each subset of features~\citep{lipovetsky2001analysis, vstrumbelj2009explaining}. These methods also vary in other respects, as we describe below.

We refer to this class of approaches as \textit{removal-based explanations} and identify 26\footnote{This total count does not include minor variations on the approaches we identified.} existing methods that rely on the feature removal principle, including several of the most widely used methods (SHAP, LIME, Meaningful Perturbations, permutation tests). We then develop a framework that shows how each method arises from various combinations of three choices: 1)~how the method removes features from the model, 2)~what model behavior the method analyzes, and 3)~how the method summarizes each feature's influence on the model. By characterizing each method in terms of three precise mathematical choices, we are able to systematize their shared elements and reveal that they rely on the same fundamental approach---feature removal.

The model explanation field has grown significantly in the past decade, and we take a broader view of the literature than existing unification theories. Our framework's flexibility lets us establish links between disparate classes of methods (e.g., computer vision-focused methods, global methods, game-theoretic methods, feature selection methods) and show that the literature is more interconnected than previously recognized. Exposing these underlying connections potentially raises questions about the degree of novelty in recent work, but we also believe that each method has the potential to offer unique advantages, either conceptually or computationally.

Through this work, we hope to empower users to reason more carefully about which tools to use, and we aim to provide researchers with new theoretical tools to build on in ongoing research. Our contributions include:

\begin{enumerate}[leftmargin=2pc]
    \item We present a framework that unifies 26 existing explanations methods. Our framework for \textbf{removal-based explanations} integrates classes of methods that were previously considered disjoint, including local and global approaches, as well as feature attribution and feature selection methods.
    
    \item We develop new mathematical tools to represent different approaches to removing features from ML models. \textit{Subset functions} provide a common representation for various feature removal techniques, revealing that this choice is interchangeable between methods.
    
    \item We generalize numerous explanation methods to express them within our framework, exposing connections that were often not apparent in the original works. In particular, for several approaches we disentangle the implicit aims of the methods from the approximations that make them usable in practice.
\end{enumerate}

We begin with background on the model explanation problem and a review of prior work (Section~\ref{sec:background}), and we then introduce our framework (Section~\ref{sec:removal_based_explanations}). The remaining
sections examine our framework in detail by showing how it encompasses existing methods. Section~\ref{sec:removal} discusses how methods remove features, Section~\ref{sec:behaviors} formalizes the model behaviors analyzed by each method, and Section~\ref{sec:explanations} describes each method's approach to summarizing each feature's influence. Finally, Section~\ref{sec:discussion} concludes and discusses future research directions.

\section{Background} \label{sec:background}

Here, we introduce the model explanation problem and briefly review existing approaches and related unification theories.

\subsection{Preliminaries}

Consider a supervised ML model $f$ that is used to predict a response variable $Y \in \mathcal{Y}$ using the input $X = (X_1, X_2, \ldots, X_d)$, where each $X_i$ represents an individual feature, such as a patient's age. We use uppercase symbols (e.g., $X$) to denote random variables and lowercase ones (e.g., $x$) to denote their values. We also use $\mathcal{X}$ to denote the domain of the full feature vector $X$ and $\mathcal{X}_i$ to denote the domain of each feature $X_i$. Finally, $x_S \equiv \{x_i : i \in S\}$ denotes a subset of features for $S \subseteq D \equiv \{1, 2, \ldots d\}$, and $\bar S \equiv D \setminus S$ represents a set's complement.

ML interpretability broadly aims to provide insight into how models make predictions. This is particularly important when $f$ is a complex model, such as a neural network or a decision forest. The most active area of research in the field is \textit{local interpretability}, which explains individual predictions, such as an individual patient diagnosis~\cite{lundberg2017unified, ribeiro2016should, sundararajan2017axiomatic}; in contrast, \textit{global interpretability} explains the model's behavior across the entire dataset~\cite{breiman2001random, covert2020understanding, owen2014sobol}. Both problems are usually addressed using \textit{feature attribution}, where a score is assigned to explain each feature's influence. However, recent work has also proposed the strategy of \textit{local feature selection}~\citep{chen2018learning}, and other papers have introduced methods to identify sets of relevant features~\citep{dabkowski2017real, fong2017interpretable, zhou2014object}.

Whether the aim is local or global interpretability, explaining the inner workings of complex models is fundamentally difficult, so it is no surprise that researchers keep devising new approaches. Commonly cited categories of approaches include perturbation-based methods~\citep{lundberg2017unified, zeiler2014visualizing}, gradient-based methods~\citep{simonyan2013deep, sundararajan2017axiomatic}, and inherently interpretable models~\citep{rudin2019stop, zhou2016learning}. However, these categories refer to loose collections of approaches that seldom share a precise mechanism.

Besides the inherently interpretable models, virtually all of these approaches generate explanations by considering some class of perturbation to the input and using the outcomes to explain each feature's influence. Certain methods consider infinitesimal perturbations by calculating gradients~\citep{simonyan2013deep, smilkov2017smoothgrad, sundararajan2017axiomatic, xu2020attribution}, but there are many possible perturbations~\citep{fong2017interpretable, lundberg2017unified, ribeiro2016should, zeiler2014visualizing}. Our work is based on the observation that numerous perturbation strategies can be understood as simulating feature removal.

\subsection{Related work}

Prior work has made solid progress in exposing connections among disparate explanation methods. Lundberg \& Lee proposed the unifying framework of \textit{additive feature attribution methods} and showed that LIME, DeepLIFT, LRP and QII are all related to SHAP~\citep{bach2015pixel, datta2016algorithmic, lundberg2017unified, ribeiro2016should, shrikumar2016not}. Similarly, Ancona et al. showed that Grad * Input, DeepLIFT, LRP and Integrated Gradients are all understandable as modified gradient backpropagations~\cite{ancona2017towards, shrikumar2016not, sundararajan2017axiomatic}. Most recently, Covert et al. showed that several global explanation methods can be viewed as \textit{additive importance measures}, including permutation tests, Shapley Net Effects, and SAGE~\citep{breiman2001random, covert2020understanding, lipovetsky2001analysis}.

Relative to prior work, the unification we propose is considerably broader but nonetheless precise. As we describe below, our framework characterizes methods along three dimensions. The choice of how to remove features has been considered by many works~\citep{aas2019explaining, frye2020shapley, hooker2019please, janzing2019feature, lundberg2017unified, merrick2019explanation, sundararajan2019many, chang2018explaining, agarwal2019explaining}. The choice of what model behavior to analyze has been considered explicitly by only a few works~\citep{covert2020understanding, lundberg2020local}, as has the choice of how to summarize each feature's influence based on a set function~\citep{covert2020understanding, datta2016algorithmic, frye2019asymmetric, lundberg2017unified, vstrumbelj2009explaining}. To our knowledge, ours is the first work to consider all three dimensions simultaneously and unite them under a single framework.

Besides the methods that we focus on, there are also methods that do not rely on the feature removal principle. We direct readers to survey articles for a broader overview of the literature~\cite{adadi2018peeking, guidotti2018survey}.

\section{Removal-Based Explanations} \label{sec:removal_based_explanations}

We now introduce our framework and briefly describe the methods it unifies.

\begin{table}[t]
\caption{Choices made by existing removal-based explanations.}
\label{tab:methods}
\begin{center}
\begin{scriptsize}
\begin{tabular}{lccc}
\toprule
\textsc{Method} & \textsc{Removal} & \textsc{Behavior} & \textsc{Summary} \\
\midrule
IME (2009) & Separate models & Prediction & Shapley value \\
IME (2010) & Marginalize (uniform) & Prediction & Shapley value \\
QII & Marginalize (marginals product) & Prediction & Shapley value \\
SHAP & Marginalize (conditional/marginal) & Prediction & Shapley value \\
KernelSHAP & Marginalize (marginal) & Prediction & Shapley value \\
TreeSHAP & Tree distribution & Prediction & Shapley value \\
LossSHAP & Marginalize (conditional) & Prediction loss & Shapley value \\
SAGE & Marginalize (conditional) & Dataset loss (label) & Shapley value \\
Shapley Net Effects & Separate models (linear) & Dataset loss (label) & Shapley value \\
SPVIM & Separate models & Dataset loss (label) & Shapley value \\
Shapley Effects & Marginalize (conditional) & Dataset loss (output) & Shapley value \\
Permutation Test & Marginalize (marginal) & Dataset loss (label) & Remove individual \\
Conditional Perm. Test & Marginalize (conditional) & Dataset loss (label) & Remove individual \\
Feature Ablation (LOCO) & Separate models & Dataset loss (label) & Remove individual \\
Univariate Predictors & Separate models & Dataset loss (label) & Include individual \\
L2X & Surrogate & Prediction loss (output) & High-value subset \\
REAL-X & Surrogate & Prediction loss (output) & High-value subset \\
INVASE & Missingness during training & Prediction mean loss & High-value subset \\
LIME (Images) & Default values & Prediction & Additive model \\
LIME (Tabular) & Marginalize (replacement dist.) & Prediction & Additive model \\
PredDiff & Marginalize (conditional) & Prediction & Remove individual \\
Occlusion & Zeros & Prediction & Remove individual \\
CXPlain & Zeros & Prediction loss & Remove individual \\
RISE & Zeros & Prediction & Mean when included \\
MM & Default values & Prediction & Partitioned subsets \\
MIR & Extend pixel values & Prediction & High-value subset \\
MP & Blurring & Prediction & Low-value subset \\
EP & Blurring & Prediction & High-value subset \\
FIDO-CA & Generative model & Prediction & High-value subset \\
\bottomrule
\end{tabular}
\end{scriptsize}
\end{center}
\vskip -0.2in
\end{table}

\subsection{A unified framework}

We develop a unified model explanation framework by connecting methods that define a feature's influence through the impact of removing it from a model. This perspective encompasses a substantial portion of the explainability literature: we find that 26 existing methods rely on this mechanism, including many of the most widely used approaches~\citep{breiman2001random, fong2017interpretable, lundberg2017unified, ribeiro2016should}.

These methods all remove groups of features from the model, but, beyond that, they take a diverse set of approaches. For example, LIME fits a linear model to an \textit{interpretable representation} of the input~\citep{ribeiro2016should}, L2X selects the most informative features for a single example~\citep{chen2018learning}, and Shapley Effects examines how much of the model's variance is explained by each feature~\citep{owen2014sobol}. Perhaps surprisingly, their differences are easy to systematize because each method removes discrete sets of features.

As our main contribution, we introduce a framework that shows how these methods can be specified using only three choices.

\begin{definition}
    {\normalfont \textbf{Removal-based explanations}} are model explanations that quantify the impact of removing sets of features from the model. These methods are determined by three choices:
    
    \begin{enumerate}
        \item (Feature removal) How the method removes features from the model (e.g., by setting them to default values or by marginalizing over a distribution of values)
        \item (Model behavior) What model behavior the method analyzes (e.g., the probability of the true class or the model loss)
        \item (Summary technique) How the method summarizes each feature's impact on the model (e.g., by removing a feature individually or by calculating the Shapley values)
    \end{enumerate}
\end{definition}

This precise yet flexible framework represents each choice as a specific type of mathematical function, as we show later. The framework unifies disparate explanation methods by unraveling each method's separate choices along these three dimensions. By allowing explicit reasoning about the trade-offs among different approaches, this perspective offers a step towards a better understanding of the literature.


\begin{figure*}[t]
\vskip -0.2in
\begin{center}
\includegraphics[trim=1.4cm 0cm 6.7cm 0cm, clip=true, width=\textwidth]{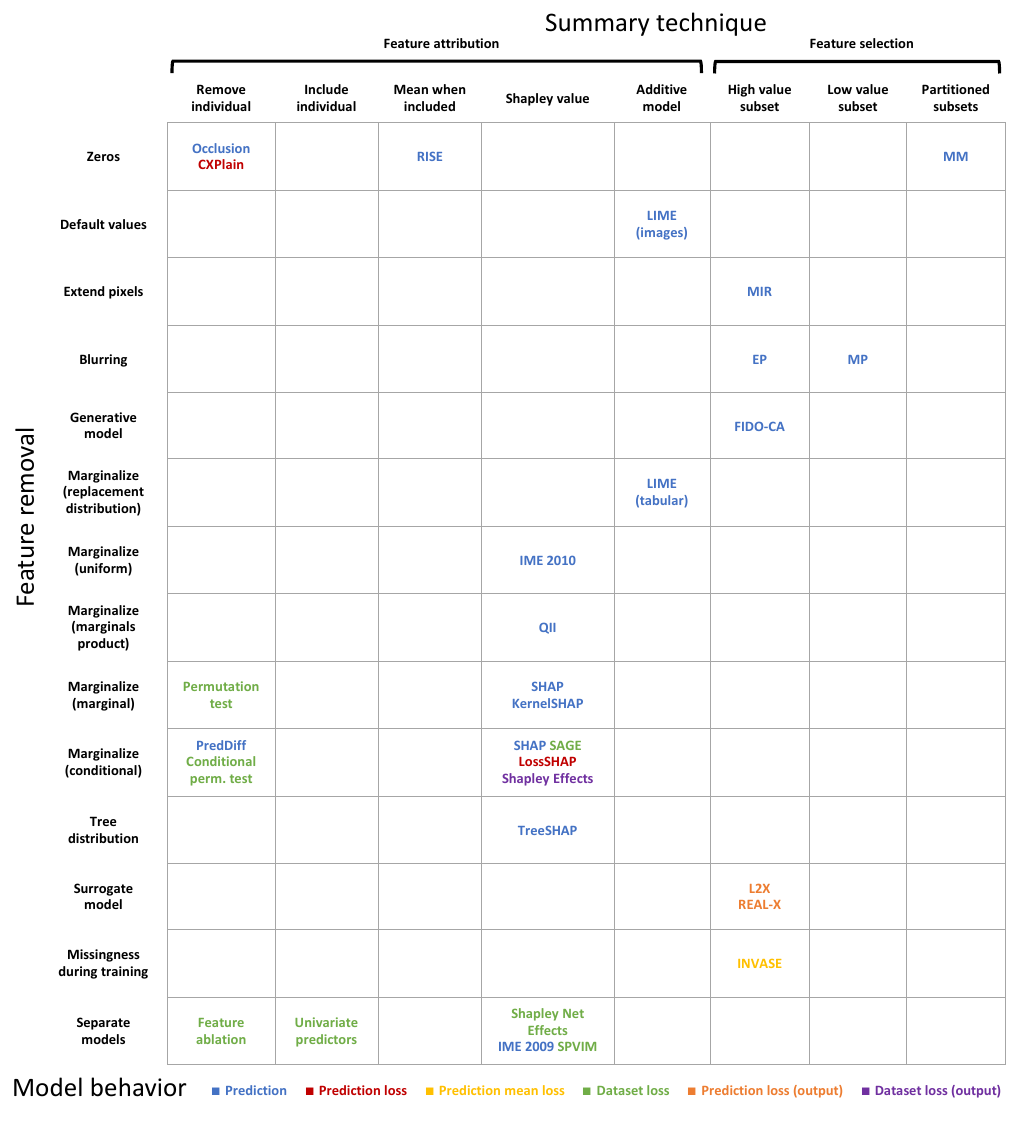}
\caption{Visual depiction of the space of removal-based explanations.}
\label{fig:methods_grid}
\end{center}
\vskip -0.2in
\end{figure*}

\subsection{Overview of existing approaches}

We now outline some of our findings, which we present in more detail in the following
sections. In particular, we preview how existing methods fit into our framework and highlight groups of methods that appear similar in light of our feature removal perspective.

Table~\ref{tab:methods} lists the methods unified by our framework
(with acronyms introduced in the next section). These methods represent diverse parts of the interpretability literature, including global methods~\citep{breiman2001random, owen2014sobol}, computer vision-focused methods~\citep{petsiuk2018rise, zeiler2014visualizing, zhou2014object, fong2017interpretable}, game-theoretic methods~\citep{covert2020understanding, lundberg2017unified, vstrumbelj2010efficient} and feature selection methods~\citep{chen2018learning, fong2019understanding, yoon2018invase}. They all have a shared
reliance on feature removal.

Disentangling the details of each method shows that many approaches share one or more of the same choices. For example, most methods choose to explain individual predictions (the model behavior), and the most popular summary technique is the Shapley value~\citep{shapley1953value}. These common choices raise important questions about how different these methods truly are and how their choices are justified.

To highlight similarities among the methods, we visually depict the space of removal-based explanations in Figure~\ref{fig:methods_grid}. Visualizing our framework reveals several regions in the space of methods that are crowded (e.g., methods that marginalize out removed features with their conditional distribution and that calculate Shapley values), while certain methods are relatively unique and spatially isolated (e.g., RISE, or LIME for tabular data).
Empty positions in the grid reveal opportunities to develop new methods; in fact, every empty position represents a viable new explanation method.
\begin{table}[t]
\caption{Common combinations of choices in existing methods. Check marks (\cmark) indicate choices that are identical between methods.}
\label{tab:common_combinations}
\begin{center}
\begin{small}
\begin{tabular}{cccc}
\toprule
\textsc{Removal} & \textsc{Behavior} & \textsc{Summary} & \textsc{Methods} \\
\midrule
& \cmark & \cmark & \makecell{IME, QII, SHAP, KernelSHAP, TreeSHAP} \\
\midrule
\cmark & & \cmark & \makecell{SHAP, LossSHAP, SAGE, Shapley Effects} \\
\midrule
\cmark & \cmark & & \makecell{Occlusion, LIME (images), MM, RISE} \\ 
\midrule
& \cmark & \cmark & \makecell{Feature ablation (LOCO), permutation tests,\\conditional permutation tests} \\
\midrule
\cmark & \cmark & & \makecell{Univariate predictors, feature ablation (LOCO),\\Shapley Net Effects, SPVIM} \\
\midrule
& \cmark & \cmark & SAGE, Shapley Net Effects, SPVIM \\
\midrule
\cmark & \cmark & & \makecell{SAGE, conditional permutation tests} \\
\midrule
\cmark & & \cmark & \makecell{Shapley Net Effects, SPVIM, IME (2009)} \\
\midrule
\cmark & & \cmark & Occlusion, CXPlain \\
\midrule
& \cmark & \cmark & Occlusion, PredDiff \\ 
\midrule
\cmark & & \cmark & \makecell{Conditional permutation tests, PredDiff} \\
\midrule
\cmark & \cmark & & SHAP, PredDiff \\
\midrule
\cmark & \cmark & & MP, EP \\
\midrule
& \cmark & \cmark & EP, FIDO-CA \\
\midrule
\cmark & \cmark & \cmark & L2X, REAL-X\footnotemark[3] \\
\midrule
\cmark & \cmark & \cmark & Shapley Net Effects, SPVIM\footnotemark[4] \\
\bottomrule
\end{tabular}
\end{small}
\end{center}
\vskip -0.2in
\end{table}

\footnotetext[3]{Although they share all three choices, L2X and REAL-X generate different explanations due to REAL-X's modified surrogate model training approach (Appendix~\ref{app:methods}).}
\addtocounter{footnote}{1}
\footnotetext[4]{SPVIM generalizes Shapley Net Effects to black-box models by using an efficient Shapley value estimation technique (Appendix~\ref{app:methods}).}
\addtocounter{footnote}{1}

Finally, Table~\ref{tab:common_combinations} shows groups of methods that differ in only one dimension of the framework. These methods are neighbors in the space of explanation methods (Figure~\ref{fig:methods_grid}), and it is remarkable how many instances of neighboring methods exist in the literature. Certain methods even have neighbors along every dimension of the framework (e.g., SHAP, SAGE, Occlusion, PredDiff, conditional permutation tests), reflecting how intimately connected the literature has become. The explainability literature is evolving and maturing, and our perspective provides a new approach for reasoning about the subtle relationships and trade-offs among existing approaches.

\section{Feature Removal} \label{sec:removal}

Here, we define the mathematical tools necessary to remove features from ML models and then examine how existing explanation methods remove features.

\subsection{Functions on subsets of features}

Most ML models make predictions given a specific set of features $X = (X_1, \ldots, X_d)$.
Mathematically, these models are functions are of the form $f: \mathcal{X} \mapsto \mathcal{Y}$, and we use $\mathcal{F}$ to denote the set of all such possible mappings. The principle behind removal-based explanations is to remove certain features to understand their impact on a model, but since most models require all the features to make predictions, removing a feature is more difficult
than simply not giving the model access to it. 

To remove features from a model, or to make predictions given a subset of features, we require a different mathematical object than $f \in \mathcal{F}$. Instead of functions with domain $\mathcal{X}$, we consider functions with domain $\mathcal{X} \times \mathcal{P}(D)$, where $\mathcal{P}(D)$ denotes the power set of $D \equiv \{1, \ldots, d\}$. To ensure invariance to the held-out features, these functions must depend only on a set of features specified by a subset $S \in \mathcal{P}(D)$,
so we formalize \textit{subset functions} as follows.

\vskip 0.2in

\begin{definition} \label{def:subset_function}
    {\normalfont A \textbf{subset function}} is a mapping of the form
    
    \begin{equation*}
        F: \mathcal{X} \times \mathcal{P}(D) \mapsto \mathcal{Y}
    \end{equation*}
    
    \noindent that is invariant to the dimensions that are not in the specified subset. That is, we have $F(x, S) = F(x', S)$ for all $(x, x', S)$ such that $x_S = x'_S$. We define $F(x_S) \equiv F(x, S)$ for convenience because the held-out values $x_{\bar S}$ are not used by $F$.
\end{definition}

A subset function's invariance property is crucial to ensure that only the specified feature values determine the function's output, while guaranteeing that the other feature values do not matter. Another way of viewing subset functions is that they simulate the use of partial inputs or missing data.
While we use $\mathcal{F}$ to represent standard prediction functions, we use $\mathfrak{F}$ to denote the set of all possible subset functions.

We introduce subset functions here because they help conceptualize how different methods remove features from ML models. Removal-based explanations typically begin with an existing model $f \in \mathcal{F}$, and in order to quantify each feature's influence, they must establish a convention for removing it from the model. A natural approach is to define a subset function $F \in \mathfrak{F}$ based on the original model $f$. To formalize this idea, we define a model's \textit{subset extension} as follows.

\begin{definition} \label{def:extension}
    A {\normalfont \textbf{subset extension}} of a model $f \in \mathcal{F}$ is a subset function $F \in \mathfrak{F}$ that agrees with $f$ in the presence of all features. That is, the model $f$ and its subset extension $F$ must satisfy
    
    \begin{equation*}
        F(x) = f(x) \quad \forall x \in \mathcal{X}.
    \end{equation*}
\end{definition}

    

As we show next, specifying a subset function $F \in \mathfrak{F}$, often as a subset extension of an existing model $f \in \mathcal{F}$, is the first step towards defining a removal-based explanation.


\subsection{Removing features from machine learning models} \label{sec:missingness_strategies}

Existing methods have devised numerous ways to evaluate models while withholding groups of features. Although certain methods use different terminology to describe their approaches (e.g., deleting information, ignoring features, using neutral values, etc.), the goal of all these methods is to measure a feature's influence through the impact of removing it from the model. Most proposed techniques can be understood as subset extensions $F \in \mathfrak{F}$ of an existing model $f \in \mathcal{F}$ (Definition~\ref{def:extension}).

We now examine each method's specific approach (see Appendix~\ref{app:methods} for more details):

\begin{itemize}[leftmargin=2pc]
    \item (Zeros) Occlusion~\citep{zeiler2014visualizing}, RISE~\citep{petsiuk2018rise} and causal explanations (CXPlain)~\citep{schwab2019cxplain} remove features simply by setting them to zero:
    
    \begin{equation}
        F(x_S) = f(x_S, 0). \label{eq:zeros}
    \end{equation}
    
    \item (Default values) LIME for image data~\citep{ribeiro2016should} and the Masking Model method (MM)~\citep{dabkowski2017real} remove features by setting them to user-defined default values (e.g., gray pixels for images). Given default values $r \in \mathcal{X}$, these methods calculate

    \begin{equation}
        F(x_S) = f(x_S, r_{\bar S}). \label{eq:default}
    \end{equation}
    
    This is a generalization of the previous approach, and in some cases features may be given different default values (e.g., their mean).
    
    
    
    
    
    \item (Extend pixel values) Minimal image representation (MIR)~\citep{zhou2014object} removes features in images by extending the values of neighboring pixels. This effect is achieved through a gradient-space manipulation.
    
    \item (Blurring) Meaningful Perturbations (MP)~\citep{fong2017interpretable} and Extremal Perturbations (EP)~\citep{fong2019understanding} remove features from images by blurring them with a Gaussian kernel. This approach is \textit{not} an extension of $f$ because the blurred image retains dependence on the removed features. Blurring fails to remove large, low frequency objects (e.g., mountains), but it provides an approximate way to remove information from images.
    
    \item (Generative model) FIDO-CA~\cite{chang2018explaining} removes feature by replacing them with a sample from a conditional generative model (e.g.~\cite{yu2018generative}). The held-out features are drawn from a generative model represented by $p_G(X_{\bar S} | X_S)$, or $\tilde x_{\bar S} \sim p_G(X_{\bar S} | X_S)$ and predictions are made as follows:
    
    \begin{equation}
        F(x_S) = f(x_S, \tilde x_{\bar S}).
    \end{equation}
    
    \item (Marginalize with conditional) SHAP~\cite{lundberg2017unified}, LossSHAP~\cite{lundberg2020local} and SAGE~\cite{covert2020understanding} present a strategy for removing features by marginalizing them out using their conditional distribution, denoted by $p(X_{\bar S} \mid X_S = x_S)$:
    
    \begin{equation}
        F(x_S) = \E\big[f(X) \mid X_S = x_S\big]. \label{eq:conditional_expectation}
    \end{equation}
    
    This approach is computationally challenging in practice, but recent works try to achieve close approximations~\cite{aas2019explaining, aas2021explaining, frye2020shapley}. Shapley Effects~\cite{owen2014sobol} implicitly uses this convention to analyze function sensitivity, while conditional permutation tests~\cite{strobl2008conditional} and Prediction Difference Analysis (PredDiff~\cite{zintgraf2017visualizing}) propose simple approximations, with the latter conditioning only on groups of bordering pixels.

    
    

    \item (Marginalize with marginal) KernelSHAP (a practical implementation of SHAP)~\cite{lundberg2017unified} removes
    features by marginalizing them out using their joint marginal distribution $p(X_{\bar S})$:
    
    \begin{equation}
        F(x_S) = \E\big[f(x_S, X_{\bar S})\big]. \label{eq:marginal}
    \end{equation}
    
    This is the default behavior in SHAP's implementation,\footnote{\url{https://github.com/slundberg/shap}} and recent work discusses the benefits of this approach~\citep{janzing2019feature}. Permutation tests~\citep{breiman2001random} also use this approach to remove individual features from a model.

    \item (Marginalize with product of marginals) Quantitative Input Influence (QII)~\citep{datta2016algorithmic} removes held-out features by marginalizing them out using the product of the marginal distributions $p(X_i)$:
    
    \begin{equation}
        F(x_S) = \E_{\prod_{i \in D} p(X_i)}\big[f(x_S, X_{\bar S})\big]. \label{eq:qii}
    \end{equation}
    
    \item (Marginalize with uniform) The updated version of the Interactions Method for Explanation (IME)~\citep{vstrumbelj2010efficient}
    removes features by marginalizing them out with a uniform distribution over the feature space. If we let $u_i(X_i)$ denote a uniform distribution over $\mathcal{X}_i$ (with extremal values defining the boundaries for continuous features), then features are removed as follows:
    
    \begin{equation}
        F(x_S) = \E_{\prod_{i \in D} u_i(X_i)}\big[f(x_S, X_{\bar S})\big]. \label{eq:ime_uniform}
    \end{equation}
    
    \item (Marginalize with replacement distributions) LIME for tabular data replaces features with independent draws from \textit{replacement distributions} (our term), each of which depends on the original feature values. When a feature $X_i$ with value $x_i$ is removed, discrete features are drawn from the distribution $p(X_i \; | \; X_i \neq x_i)$; when quantization is used for continuous features (LIME's default behavior\footnote{\url{https://github.com/marcotcr/lime}}), continuous features are simulated by first generating a different quantile and then simulating from a truncated normal distribution within that bin. If we denote each feature's replacement distribution given the original value $x_i$ as $q_{x_i}(X_i)$, then LIME for tabular data removes features as follows:
    
    \begin{equation}
        F(x, S) = \E_{\prod_{i \in D} q_{x_i}(X_i)}\big[f(x_S, X_{\bar S})\big].
    \end{equation}
    
    Although this function $F$ agrees with $f$ given all features, it is \textit{not} an extension because it does not satisfy the invariance property for subset functions.
    
    \item (Tree distribution) Path-dependent TreeSHAP~\cite{lundberg2020local} removes features using the distribution induced by the underlying tree model, which roughly approximates the conditional distribution. When splits for held-out
    features are encountered in the model's trees, TreeSHAP averages predictions from the multiple paths in proportion to how often the dataset follows each path.
    
    \item (Surrogate model) Learning to Explain (L2X~\cite{chen2018learning}) and REAL-X~\cite{jethani2021have} train separate surrogate models $F$ to match the original model's predictions when groups of features are held out. The surrogate model accommodates missing features, allowing us to represent it as a subset function $F \in \mathfrak{F}$, and it aims to provide the following approximation:
    
    \begin{equation}
        F(x_S) \approx \E\big[f(X) \mid X_S = x_S\big]. \label{eq:surrogate}
    \end{equation}
    
    The surrogate model approach was also proposed separately in the context of Shapley values~\cite{frye2020shapley}.
    
    \item (Missingness during training) Instance-wise Variable Selection (INVASE~\cite{yoon2018invase}) uses a model that has missingness introduced at training time. Removed features are replaced with zeros, so that the model makes the following approximation:
    
    \begin{equation}
        F(x_S) = f(x_S, 0) \approx p(Y \mid X_S = x_S). \label{eq:missingness}
    \end{equation}
    
    This approximation occurs for models trained cross entropy loss, but other loss functions may lead to different results (e.g., the conditional expectation for MSE loss). Introducing missingness during training differs from the default values approach because the model is trained to recognize zeros (or other replacement values) as missing values rather than zero-valued features.
    
    \item (Separate models) The original version of IME~\cite{vstrumbelj2009explaining} is not based on a single model $f$, but rather on separate models trained for each feature subset, or $\{f_S : S \subseteq D\}$. The prediction for a subset of features is given by that subset's model:
    
    \begin{equation}
        F(x_S) = f_S(x_S).
    \end{equation}
    
    Shapley Net Effects~\cite{lipovetsky2001analysis} uses an identical approach in the context of linear models, with SPVIM generalizing the approach to black-box models~\cite{williamson2020efficient}. Similarly, feature ablation, also known as leave-one-covariate-out (LOCO~\cite{lei2018distribution}), trains models to remove individual features, and the univariate predictors approach (used mainly for feature selection) uses models trained with individual features~\cite{guyon2003introduction}.
    Although the separate models approach is technically a subset extension of the model $f_D$ trained with all features, its predictions given subsets of features are not based on $f_D$.
    
    
    
    
\end{itemize}

Most of these approaches are subset extensions of an existing model $f$, so our formalisms provide useful tools for understanding how removal-based explanations remove features from models. However, consider two exceptions: the blurring technique (MP and EP) and LIME's approach with tabular data. Both provide functions of the form $F: \mathcal{X} \times \mathcal{P}(D) \mapsto \mathcal{Y}$ that agree with $f$ given all features, but that still exhibit dependence on removed features. Based on our mathematical characterization of subset functions and their invariance to held-out features, we argue that these two approaches do not fully remove features from the model. We conclude that the first dimension of our framework amounts to choosing an extension $F \in \mathfrak{F}$ of the model $f \in \mathcal{F}$.


\section{Explaining Different Model Behaviors} \label{sec:behaviors}

Removal-based explanations all aim to demonstrate how a model works, but they can do so by analyzing a variety of model behaviors. We now consider the various choices of target quantities to observe as different features are withheld from the model.

The feature removal principle is flexible enough to explain virtually any function. For example, methods can explain a model's prediction, a model's loss function, a hidden layer in a neural network, or any node in a computation graph. In fact, removal-based explanations need not be restricted to the ML context: any function that accommodates missing inputs can be explained via feature removal by examining either its output or some function of its output as groups of inputs are removed. This perspective shows the broad potential applications for removal-based explanations.

However, since our focus is the ML context, we proceed by examining
how existing model explanation methods work. Each method's target quantity can be understood as a function of the model output, which is represented by a subset function $F(x_S)$. Many methods explain the model output or a simple function of the output, such as the log-odds ratio. Other methods
take into account a measure of the model's loss, for either an individual input or the entire dataset. Ultimately, as we show below, each method generates explanations based on a set function of the form

\begin{equation*}
    u: \mathcal{P}(D) \mapsto \R,
\end{equation*}

\noindent which represents a value associated with each subset of features $S \subseteq D$. This set function represents the model behavior that a method is designed to explain.

We now examine the specific choices made by existing methods (see Appendix~\ref{app:methods} for further details on each method).
The various model behaviors that methods analyze, and their corresponding set functions, include:

\begin{itemize}[leftmargin=2pc]
    \item (Prediction) Occlusion, RISE, PredDiff, MP, EP, MM, FIDO-CA, MIR, LIME, SHAP (including KernelSHAP and TreeSHAP), IME and QII all analyze a model's prediction for an individual input $x \in \mathcal{X}$:
    
    \begin{equation}
        u_x(S) = F(x_S). \label{eq:output_game}
    \end{equation}
    
    These methods examine how holding out different features makes an individual prediction either higher or lower. For multi-class classification models, methods often use a single output that corresponds to the class of interest, and they can optionally apply a simple function to the model's output (for example, using the log-odds ratio rather than the classification probability).
    
    \item (Prediction loss) LossSHAP and CXPlain take into account the true label $y$ for an input $x$ and calculate the prediction loss using a loss function $\ell$:
    
    \begin{equation}
        v_{xy}(S) = - \ell\big(F(x_S), y\big). \label{eq:shap_loss_game}
    \end{equation}
    
    By incorporating the label, these methods quantify whether certain features make the prediction more or less correct. Note that the minus sign is necessary to give the set function a higher value when more informative features are included.
    
    \item (Prediction mean loss) INVASE considers the expected loss for a given input $x$ according to the label's conditional distribution $p(Y \mid X = x)$:
    
    \begin{equation}
        v_x(S) = - \E_{p(Y \mid X = x)}\Big[\ell\big(F(x_S), Y\big)\Big]. \label{eq:pml_game}
    \end{equation}
    
    By averaging the loss across the label's distribution, INVASE highlights features that correctly predict what \textit{could} have occurred, on average.
    
    \item (Dataset loss) Shapley Net Effects, SAGE, SPVIM, feature ablation, permutation tests and univariate predictors consider the expected loss across the entire dataset:
    
    \begin{equation}
        v(S) = - \E_{XY}\Big[\ell\big(F(X_S), Y\big)\Big]. \label{eq:expected_loss_game}
    \end{equation}
    
    These methods quantify how much the model's performance degrades when different features are removed. This set function can also be viewed as the predictive power derived from sets of features~\citep{covert2020understanding}, and recent work has proposed a SHAP value aggregation that is a special case of this approach~\citep{frye2020shapley}.
    
    \item (Prediction loss w.r.t. output) L2X and REAL-X consider the loss between the full model output and the prediction given a subset of features:
    
    \begin{equation}
        w_x(S) = - \ell\big(F(x_S), F(x)\big). \label{eq:plo_game}
    \end{equation}
    
    These methods highlight features that on their own lead to similar predictions as the full feature set.
    
    \item (Dataset loss w.r.t. output) Shapley Effects considers the expected loss with respect to the full model output:
    
    \begin{equation}
        w(S) = - \E_X\Big[\ell\big(F(X_S), F(X)\big)\Big]. \label{eq:shapley_effects_game}
    \end{equation}
    
    Though related to the dataset loss approach~\citep{covert2020understanding}, this approach focuses on each feature's influence on the model output rather than on the model performance.
\end{itemize}

Each set function serves a distinct purpose in exposing a model's dependence on different features. The first three approaches listed above analyze the model's behavior for individual predictions (local explanations) while the last two take into account the model's behavior across the entire dataset (global explanations). Although their aims differ, these set functions are all in fact related. Each builds upon the previous ones by accounting for either the loss or data distribution, and their relationships can be summarized as follows:

\begin{align}
    v_{xy}(S) &= - \ell\big(u_x(S), y\big) \\
    w_x(S) &= - \ell\big( u_x(S), u_x(D) \big) \\
    v_x(S) &= \E_{p(Y \mid X = x)}\big[v_{xY}(S)\big] \\
    v(S) &= \E_{XY}\big[v_{XY}(S)\big] \label{eq:loss_shap_sage} \\
    w(S) &= \E_X\big[w_X(S)\big]
\end{align}






These relationships show that explanations based on one set function are in some cases related to explanations based on another. For example, Covert et al. showed that SAGE explanations are the expectation of explanations provided by LossSHAP~\cite{covert2020understanding}---a relationship reflected in Eq.~\ref{eq:loss_shap_sage}.

Understanding these connections is possible only because our framework disentangles each method's choices rather than viewing each method as a monolithic algorithm. We conclude by reiterating that removal-based explanations can explain virtually any function, and that choosing what to explain amounts to selecting a set function $u: \mathcal{P}(D) \mapsto \R$ to represent the model's dependence
on different sets of features.

\section{Summarizing Feature Influence} \label{sec:explanations}

The third choice for removal-based explanations is how to summarize each feature's influence on the model. We examine the various summarization techniques and then discuss their computational complexity and approximation approaches.

\subsection{Explaining set functions} \label{sec:summarization_strategies}

The set functions we used to represent a model's dependence on different features
(Section~\ref{sec:behaviors}) are complicated mathematical objects that are difficult to communicate fully due to the exponential number of feature subsets and underlying feature interactions. Removal-based explanations confront this challenge by providing users with a concise summary of each feature's influence.

We distinguish between two main types of summarization approaches: feature attributions and feature selections. Many methods provide explanations in the form of \textit{feature attributions}, which are numerical scores $a_i\in \R$ given to each feature $i = 1, \ldots, d$. If we use $\mathcal{U}$ to denote the set of all functions $u: \mathcal{P}(D) \mapsto \R$, then we can represent feature attributions as mappings of the form $E: \mathcal{U} \mapsto \R^d$, which we refer to as \textit{explanation mappings}. Other methods take the alternative approach of summarizing set functions with a set $S^* \subseteq D$ of the most influential features. We represent these \textit{feature selection} summaries as explanation mappings of the form $E: \mathcal{U} \mapsto \mathcal{P}(D)$. Both approaches provide users with simple summaries of a feature's contribution to the set function.

We now consider the specific choices made by each method (see Appendix~\ref{app:methods} for further details). For simplicity, we let $u$ denote the set function each method analyzes. Surveying the various removal-based explanation methods, the techniques for summarizing each feature's influence include:

\begin{itemize}[leftmargin=2pc]
    \item (Remove individual) Occlusion, PredDiff, CXPlain, permutation tests and feature ablation (LOCO) calculate the impact of removing a single feature from the model, resulting in the following attribution values:
    
    \begin{equation}
        a_i= u(D) - u(D \setminus \{i\}). \label{eq:remove_individual}
    \end{equation}
    
    Occlusion, PredDiff and CXPlain can also be applied with groups of features, or superpixels, in image contexts.
    
    
    \item (Include individual) The univariate predictors approach calculates the impact of including individual features, resulting in the following attribution values:
    
    \begin{equation}
        a_i= u(\{i\}) - u(\{\}). \label{eq:include_individual}
    \end{equation}
    
    This is essentially the reverse of the previous approach: rather than removing individual features from the complete set, this approach adds individual features to the empty set.
    
    \item (Additive model) LIME fits a regularized additive model to a dataset of perturbed examples. In the limit of an infinite number of samples, this process approximates the following attribution values:
    
    \begin{equation}
        a_1, \ldots, a_d = \argmin_{b_0, \ldots, b_d} \; \sum_{S \subseteq D}\pi(S)\Big(b_0 + \sum_{i \in S} b_i - u(S)\Big)^2 + \Omega(b_1, \ldots, b_d). \label{eq:lime}
    \end{equation}
    
    In this problem, $\pi$ represents a weighting kernel and $\Omega$ is a regularization function that is often set to the $\ell_1$ penalty to encourage sparse attributions~\citep{tibshirani1996regression}. Since this summary is based on an additive model, the learned coefficients $(a_1, \ldots, a_d)$ represent the incremental value associated with including each feature.
    
    \item (Mean when included) RISE determines feature attributions by sampling many subsets $S \subseteq D$ and then calculating the mean value when a feature is included. Denoting the distribution of subsets as $p(S)$ and the conditional distribution as $p(S \mid i \in S)$, the attribution values are defined as
    
    \begin{equation}
        a_i= \E_{p(S \mid i \in S)}\big[u(S)\big].
    \end{equation}
    
    In practice, RISE samples the subsets $S \subseteq D$ by removing each feature $i$ independently with probability $p$, using $p = 0.5$ in their experiments~\citep{petsiuk2018rise}.
    
    \item (Shapley value) Shapley Net Effects, IME, Shapley Effects, QII, SHAP (including KernelSHAP, TreeSHAP and LossSHAP), SPVIM and SAGE all calculate feature attributions using the Shapley value, which we denote as $a_i = \phi_i(u)$.
    Shapley values are the only attributions that satisfy several desirable properties~\cite{shapley1953value}, and they are defined as follows:
    
    \begin{equation}
        \phi_i(v) = \frac{1}{d} \sum_{S \subseteq D \setminus \{i\}} \binom{d - 1}{|S|}^{-1} \Big( v(S \cup \{i\}) - v(S) \Big).
    \end{equation}
    
    \item (Low-value subset) MP selects a small set of features $S^*$ that can be removed to give the set function a low value. It does so by solving the following optimization problem:
    
    \begin{equation}
        S^* = \argmin_{S} \; u(D \setminus S) + \lambda |S|. \label{eq:mp_opt}
    \end{equation}
    
    In practice, MP incorporates additional regularizers and solves a relaxed version of this problem (see Section~\ref{sec:complexity}).
    
    \item (High-value subset) MIR solves an optimization problem to select a small set of features $S^*$ that alone can give the set function a high value. For a user-defined minimum value $t$, the problem is given by:

    \begin{equation}
        S^* = \argmin_S \; |S| \quad {\text s.t.} \;\; u(S) \geq t. \label{eq:mir_opt}
    \end{equation}
    
    L2X and EP solve a similar problem but switch the terms in the constraint and optimization objective. For a user-defined subset size $k$, the optimization problem is given by:
    
    \begin{equation}
        S^* = \argmax_S \; u(S) \quad \mathrm{s.t.} \;\; |S| = k. \label{eq:l2x_opt}
    \end{equation}
    
    Finally, INVASE, REAL-X and FIDO-CA solve a regularized version of the problem with a parameter $\lambda > 0$ controlling the trade-off between the subset value and subset size:
    
    \begin{equation}
        S^* = \argmax_S \; u(S) - \lambda |S|. \label{eq:invase_opt}
    \end{equation}
    
    \item (Partitioned subsets) MM solves an optimization problem to partition the features into $S^*$ and $D \setminus S^*$ while maximizing the difference in the set function's values. This approach is based on the idea that removing features to find a low-value subset (as in MP) and retaining features to get a high-value subset (as in MIR, L2X, EP, INVASE, REAL-X and FIDO-CA) are both reasonable approaches for identifying influential features. The problem is given by:
    
    \begin{equation}
        S^* = \argmax_S \; u(S) - \gamma u(D \setminus S) - \lambda |S|. \label{eq:mm_opt}
    \end{equation}
    
    In practice, MM also incorporates regularizers and monotonic link functions to enable a more flexible trade-off between $u(S)$ and $u(D \setminus S)$ (see Appendix~\ref{app:methods}).
\end{itemize}

As this discussion shows, every removal-based explanation generates summaries of each feature's influence on the underlying set function. In general, a model's dependencies are
too complex to communicate fully, so explanations must provide users with a concise summary instead. As noted, most methods we discuss generate feature attributions, but several others
generate explanations by selecting the most important features. These feature selection explanations are essentially coarse attributions that assign binary importance rather than a real number.

Interestingly, if the high-value subset optimization problems solved by MIR, L2X, EP, INVASE and FIDO-CA were applied to the set function that represents the dataset loss (Eq.~\ref{eq:loss_shap_sage}), they would resemble conventional global feature selection~\citep{guyon2003introduction}. The problem in Eq.~\ref{eq:l2x_opt} determines the set of $k$ features with maximum predictive power, the problem in Eq.~\ref{eq:mir_opt} determines the smallest possible set of features that achieve the performance represented by $t$, and the problem in Eq.~\ref{eq:invase_opt} uses a parameter $\lambda$ to control the trade-off. Though not generally viewed as a model explanation approach, global feature selection serves an identical purpose of identifying highly predictive features.

\begin{figure*}[t]
\begin{center}
\includegraphics[width=\columnwidth]{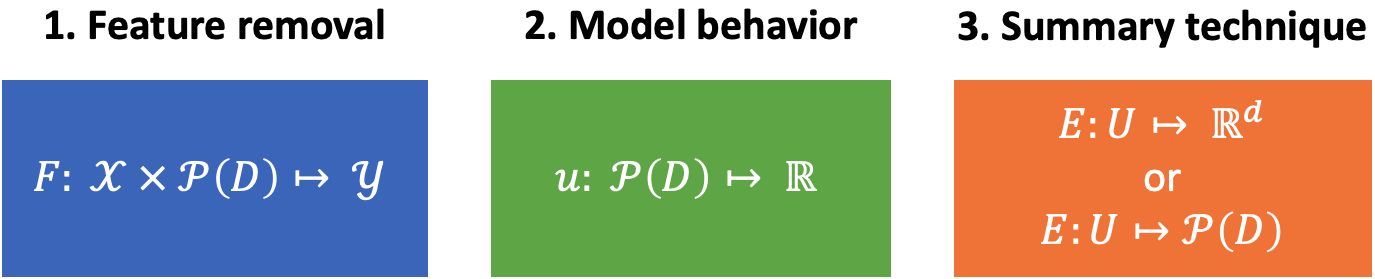}
\vskip 0.1in
\caption{Removal-based explanations are specified by three precise mathematical choices: a subset function $F \in \mathfrak{F}$, a set function $u \in \mathcal{U}$, and an explanation mapping $E$ (for feature attribution or selection).}
\label{fig:framework}
\end{center}
\end{figure*}

We conclude by reiterating that the third dimension of our framework amounts to a choice of explanation mapping, which takes the form $E : \mathcal{U} \mapsto \R^d$ for feature attribution or $E: \mathcal{U} \mapsto \mathcal{P}(D)$ for feature selection. Our discussion so far has shown that removal-based explanations can be specified using three precise mathematical choices, as depicted in Figure~\ref{fig:framework}. These methods, which are often presented in ways that make their connections difficult to discern, are constructed in a remarkably similar fashion.

\subsection{Complexity and approximations} \label{sec:complexity}

Showing how certain explanation methods fit into our framework requires distinguishing between their implicit aims
and the approximations that make them practical. Our presentation of these methods deviates from the original papers, which often focus on details of a method's implementation. We now bridge the gap by describing these methods'
computational complexity and the approximations they use out of necessity.

The challenge with most summarization techniques described above is that they require calculating the underlying set function's value $u(S)$ for many subsets of features. In fact, without making any simplifying assumptions about the model or data distribution, several techniques must examine all $2^d$ subsets of features. This includes the Shapley value, RISE's summary technique and LIME's linear model. Finding exact solutions to several of the optimization problems (MP, MIR, MM, INVASE, FIDO-CA) also requires examining all subsets of features, and solving the constrained optimization problem (EP, L2X) for $k$ features requires examining $\binom{d}{k}$ subsets, or $2^d d^{-\frac{1}{2}}$ subsets in the worst case.\footnote{This can be seen by applying Stirling's approximation to $\binom{d}{d/2}$ as $d$ becomes large.}

The only approaches with lower computational complexity are those that remove individual features (Occlusion, PredDiff, CXPlain, permutation tests, feature ablation) or include individual features (univariate predictors). These require only one subset per feature, or $d$ total feature subsets.

Many summarization techniques have superpolynomial complexity in $d$, making them intractable for large numbers of features. However, these methods work in practice due to fast approximation approaches, and in some cases methods have even been devised to generate explanations in real-time. Strategies that yield fast approximations include:

\begin{itemize}[leftmargin=2pc]
    \item Attribution values that are the expectation of a random variable can be estimated by Monte Carlo approximation. IME~\citep{vstrumbelj2010efficient}, Shapley Effects~\citep{song2016shapley} and SAGE~\citep{covert2020understanding} use sampling strategies to approximate Shapley values, and RISE also estimates its attributions via sampling~\citep{petsiuk2018rise}.
    
    \item KernelSHAP, LIME and SPVIM are based on linear regression models fitted to datasets containing an exponential number of datapoints. In practice, these techniques fit models to smaller sampled datasets, which means optimizing an approximate version of their objective function~\citep{lundberg2017unified, covert2021improving}.
    
    \item TreeSHAP calculates Shapley values in polynomial time using a dynamic programming algorithm that exploits the structure of tree-based models. Similarly, L-Shapley and C-Shapley exploit the properties of models for structured data to provide fast Shapley value approximations~\citep{chen2018shapley}.
    
    \item Several of the feature selection methods (MP, L2X, REAL-X, EP, MM, FIDO-CA) solve continuous relaxations of their discrete optimization problems. While these optimization problems can be solved by representing the set of features $S \subseteq D$ as a mask $m \in \{0, 1\}^d$, these methods instead use a continuous mask variable of the form $m \in [0, 1]^d$. When these methods incorporate a penalty on the subset size $|S|$, they also sometimes use the convex relaxation $||m||_1$.
    
    \item One feature selection method (MIR) uses a greedy optimization algorithm. MIR determines a set of influential features $S \subseteq D$ by iteratively removing groups of features that do not reduce the predicted probability for the correct class.
    
    \item One feature attribution method (CXPlain) and several feature selection methods (L2X, INVASE, REAL-X, MM) generate real-time explanations by learning separate explainer models. CXPlain learns an explainer model using a dataset consisting of manually calculated explanations, which removes the need to iterate over each feature when generating new explanations. L2X learns a model that outputs a set of features (represented by a $k$-hot vector) and INVASE/REAL-X learn similar selector models that can output arbitrary numbers of features. Similarly, MM learns a model that outputs masks of the form $m \in [0, 1]^d$ for images. These techniques can be viewed as \textit{amortized} approaches because they learn models that perform the summarization step in a single forward pass.
\end{itemize}

In conclusion, many methods have developed approximations that enable efficient model explanation, despite sometimes using summarization techniques that are inherently intractable (e.g., Shapley values). Certain techniques are considerably faster than others (i.e., the amortized approaches), and some can trade off computational cost for approximation accuracy~\citep{vstrumbelj2010efficient, covert2021improving}, but they are all sufficiently fast to be used in practice.

We speculate, however, that more approaches will be made to run in real-time by learning separate explainer models, as in the MM, L2X, INVASE, CXPlain and REAL-X approaches~\citep{dabkowski2017real, chen2018learning, yoon2018invase, schwab2019cxplain, jethani2021have}. Besides these methods, others have been proposed that learn the explanation process either as a component of the original model~\citep{fan2017adversarial, taghanaki2019infomask} or as a separate model after training~\citep{schulz2020restricting}. Such approaches may be necessary to bypass the need for multiple model evaluations and make removal-based explanations as fast as gradient-based and propagation-based methods.

\section{Discussion} \label{sec:discussion}

In this work, we developed a unified framework that characterizes a significant portion of the model explanation literature (26 existing methods). Removal-based explanations have a great degree of flexibility, and we systematized their differences by showing that each method is specified by three precise mathematical choices:

\begin{enumerate}[leftmargin=2pc]
    \item \textbf{How the method removes features.} Each method specifies a subset function $F \in \mathfrak{F}$ to make predictions with subsets of features, often based on an existing model $f \in \mathcal{F}$.
    
    \item \textbf{What model behavior the method analyzes.} Each method implicitly relies on a set function $u: \mathcal{P}(D) \mapsto \R$ to represent the model's dependence on different groups of features. The set function describes the model's behavior either for an individual prediction or across the entire dataset.

    \item \textbf{How the method summarizes each feature's influence.} Methods generate explanations that provide a concise summary of each feature's contribution to the set function $u \in \mathcal{U}$. Mappings of the form $E: \mathcal{U} \mapsto \R^d$ generate feature attribution explanations, and mappings of the form $E: \mathcal{U} \mapsto \mathcal{P}(D)$ generate feature selection explanations.
\end{enumerate}

The growing interest in black-box ML models has spurred a remarkable amount of model explanation research, and in the past decade we have seen a number of publications proposing innovative new methods. However, as the field has
matured we have also seen a growing number of unifying theories that reveal underlying similarities and implicit relationships~\cite{ancona2017towards, covert2020understanding, lundberg2017unified}. Our framework for removal-based explanations is perhaps the broadest unifying theory yet, and it bridges the gap between disparate parts of the explainability literature.

An improved understanding of the field presents new opportunities for both explainability users and researchers. For users, we hope that our framework will allow for more explicit reasoning about the trade-offs between available explanation tools.
The unique advantages of different methods are difficult to understand when they are viewed as monolithic algorithms, but disentangling their choices makes it simpler to reason about their strengths and weaknesses.

For researchers, our framework offers several promising directions for future work. We identify three key areas that can be explored to better understand the trade-offs between different removal-based explanations:

\begin{itemize}[leftmargin=2pc]
    \item Several of the methods characterized by our framework can be interpreted using ideas from information theory~\cite{chen2018learning, covert2020understanding}. We suspect that other methods can be understood with an information-theoretic perspective and that this may shed light on whether there are theoretically justified choices for each dimension of our framework.
    
    \item As we showed in Section~\ref{sec:behaviors}, every removal-based explanation is based on an underlying set function that represents the model's behavior. Set functions can be viewed as \textit{cooperative games}, and we suspect that methods besides those that use Shapley values~\cite{covert2020understanding, datta2016algorithmic, lundberg2017unified, owen2014sobol, vstrumbelj2009explaining} can be related to techniques from cooperative game theory.
    
    \item Finally, it is remarkable that so many researchers have developed, with some degree of independence, explanation methods based on the same feature removal principle. We speculate that cognitive psychology may shed light on why this represents a natural approach to explaining complex decision processes. This would be impactful for the field because, as recent work has pointed out, explainability research is surprisingly disconnected from the social sciences~\cite{miller2019explanation, miller2017explainable}.
\end{itemize}

In conclusion, as the field evolves and the number of removal-based explanations continues to grow, we hope that our framework can serve as a foundation upon which future research can build.

\section*{Acknowledgements}

We thank members of the Lee Lab for helpful discussions. This work was funded by NSF DBI-1552309 and DBI-1759487, NIH R35-GM-128638 and R01-NIA-AG-061132.

\appendix
\section{Method Details} \label{app:methods}

Here, we provide additional details about some of the explanation methods discussed in the main text. In several cases, we presented generalized versions of methods that deviated from their explanations in the original papers.

\subsection{Meaningful Perturbations (MP)}

Meaningful Perturbations \citep{fong2017interpretable} considers multiple ways of deleting information from an input image, and the approach it recommends is a blurring operation. Given a mask $m \in [0, 1]^d$, MP uses a function $\Phi(x, m)$ to denote the modified input and suggests that the mask may be used to 1) set pixels to a constant value, 2) replace them with Gaussian noise, or 3) blur the image. In the blurring approach, each pixel $x_i$ is blurred separately using a Gaussian kernel with standard deviation given by $\sigma \cdot m_i$ (for a user specified $\sigma > 0$).

To prevent adversarial solutions, MP incorporates a total variation norm on the mask, upsamples it from a low-resolution version, and uses a random jitter on the image during optimization. Additionally, MP uses a continuous mask $m \in [0, 1]^d$ in place of a binary mask $\{0, 1\}^d$ and the $\ell_1$ penalty on the mask in place of the $\ell_0$ penalty. Although MP's optimization tricks are key to providing visually compelling explanations, our presentation focuses on the most essential part of the optimization objective, which is reducing the classification probability while blurring only a small part of the image (Eq.~\ref{eq:mp_opt}).

\subsection{Extremal Perturbations (EP)}

Extremal Perturbations \citep{fong2019understanding} is an extension of MP with several modifications. The first is switching the objective from a ``removal game'' to a ``preservation game,'' which means learning a mask that retains rather than removes the salient information. The second is replacing the penalty on the subset size (or the mask norm) with a constraint. In practice, the constraint is enforced using a penalty, but the authors argue that it should still be viewed as a constraint due to the use of a large regularization parameter.

EP uses the same blurring operation as MP and introduces new tricks to ensure a smooth mask, but our presentation focuses on the most important part of the optimization problem, which is maximizing the classification probability while blurring a fixed portion of the image (Eq.~\ref{eq:l2x_opt}).

\subsection{FIDO-CA}

FIDO-CA~\cite{chang2018explaining} is similar to EP but it replaces the blurring operation with features drawn from a generative model. The generative model $p_G$ can condition on arbitrary subsets of features, and although its samples are non-deterministic, FIDO-CA achieves strong results using a single sample. The authors consider multiple generative models but recommend a generative adversarial network (GAN) that uses contextual attention~\cite{yu2018generative}. The optimization objective is based on the same ``preservation game'' as EP, and the authors use the Concrete reparameterization trick~\cite{maddison2016concrete} for optimization.

\subsection{Minimal Image Representation (MIR)}

The Minimal Image Representation approach \cite{zhou2014object} removes information from an image to determine which regions are salient for the desired class. MIR works by creating a segmentation of edges and regions and iteratively removing segments from the image (selecting those that least decrease the classification probability) until the remaining image is incorrectly classified. We view this as a greedy approach for solving the constrained optimization problem

\begin{equation*}
    \min_S \; |S| \quad \mathrm{s.t.} \;\; u(S) \geq t,
\end{equation*}

\noindent where $u(S)$ represents the prediction with the specified subset of features and $t$ represents the minimum allowable classification probability. Our presentation of MIR in the main text focuses on this view of the optimization objective rather than the specific greedy algorithm MIR uses (Eq.~\ref{eq:mir_opt}).

\subsection{Masking Model (MM)}

The Masking Model approach \citep{dabkowski2017real} observes that removing salient information (while preserving irrelevant information) and removing irrelevant information (while preserving salient information) are both reasonable approaches to understanding image classifiers. The authors refer to these tasks as discovering the smallest destroying region (SDR) and smallest sufficient region (SSR).

The authors adopt notation similar to MP \cite{fong2017interpretable}, using $\Phi(x, m)$ to denote the transformation to the input given a mask $m \in [0, 1]^d$. For an input $x \in \mathcal{X}$, the authors aim to solve the following optimization problem:

\begin{equation*}
    \min_m \; \lambda_1 \mathrm{TV}(m) + \lambda_2 ||m||_1 - \log f\big(\Phi(x, m)\big) + \lambda_3 f\big(\Phi(x, 1 - m)\big)^{\lambda_4}.
\end{equation*}

The $\mathrm{TV}$ (total variation) and $\ell_1$ penalty terms are both similar to MP and respectively encourage smoothness and sparsity in the mask. Unlike MP, MM learns a global explainer model that outputs approximate solutions to this problem in a single forward pass. In the main text, we provide a simplified presentation of the problem that does not include the logarithm in the third term or the exponent in the fourth term (Eq.~\ref{eq:mm_opt}). We view these as monotonic link functions that provide a more complex trade-off between the objectives but that are not necessary for finding informative solutions.





\subsection{Learning to Explain (L2X)}

The L2X method performs instance-wise feature selection by learning an auxiliary model $g_\alpha$ and a selector model $V_\theta$ (see Eq.~6 of Chen et al.~\cite{chen2018learning}). These models are learned jointly and are optimized via the similarity between predictions from $g_\alpha$ and from the original model, denoted as $\mathbbm{P}_m$ by \cite{chen2018learning}. With slightly modified notation that highlights the selector model's dependence on $X$, the L2X objective can be written as:

\begin{equation}
    \max_{\alpha, \theta} \; \E_{X, \zeta} \E_{Y \sim \mathbbm{P}_m(X)} \Big[ \log g_\alpha\big(V_\theta(X, \zeta) \odot X, Y\big) \Big]. \label{eq:l2x_objective}
\end{equation}

In Eq.~\ref{eq:l2x_objective}, the random variables $X$ and $\zeta$ are sampled independently, $Y$ is sampled from the model's distribution $\mathbbm{P}_m(X)$, $V_\theta(X, \zeta) \odot X$ represents an element-wise multiplication with (approximately) binary indicator variables $V_\theta(X, \zeta)$ sampled from the Concrete distribution \citep{maddison2016concrete}, and $\log g_\alpha(\cdot, Y)$ represents the model's estimate of $Y$'s log-likelihood.

We can gain more insight into this objective function by reformulating it. If we let $V_\theta(X, \zeta)$ be a deterministic function $\epsilon(X)$, interpret the log-likelihood as a loss function $\ell$ for the prediction from $g_\alpha$ (e.g., cross entropy loss) and represent $g_\alpha$ as a subset function $F$, then we can rewrite the L2X objective as follows:

\begin{equation*}
    \max_{F, \epsilon} \; \E_{X} \E_{Y \sim \mathbbm{P}_m(X)} \Big[ \ell\big( F\big(X, \epsilon(X)\big), Y\big) \Big].
\end{equation*}

Next, rather than considering the expected loss for labels $Y$ distributed according to $\mathbbm{P}_m(X)$, we can rewrite this as a loss between the subset function's prediction $F\big(X, \epsilon(X)\big)$ and the full model prediction $f(X) \equiv \mathbbm{P}_m(X)$:

\begin{equation*}
    \max_{F, \epsilon} \; \E_{X} \Big[ \ell\big( F\big(X, \epsilon(X)\big), f(X)\big) \Big].
\end{equation*}

Finally, we can see that L2X implicitly trains a surrogate model $F$ to match the original model's predictions, and that the optimization objective for each input $x \in \mathcal{X}$ is given by

\begin{equation*}
    S^* = \argmax_{|S| = k} \; \ell\big(F(x_S), f(x)\big).
\end{equation*}

This matches the description of L2X provided in the main text (Eqs.~\ref{eq:surrogate}, \ref{eq:plo_game}, \ref{eq:l2x_opt}). It is only when we have $f(x) = \mathbbm{P}_m(x) = p(Y \mid X = x)$ and $F(x_S) = p(Y \mid X_S = x_S)$ that L2X's information-theoretic interpretation holds, at least in the classification case. Or, in the regression case, where we can replace the log-likelihood with a simpler MSE loss, L2X can be interpreted in terms of conditional variance minimization (rather than mutual information maximization) when we have $f(x) = \E[Y \mid X = x]$ and $F(x_S) = \E[Y \mid X_S = x_S]$.



\subsection{Instance-wise Variable Selection (INVASE)} \label{app:invase}

INVASE \citep{yoon2018invase} is similar to L2X in that it performs instance-wise feature selection using a learned selector model. However, INVASE has several differences in its implementation and objective function. INVASE relies on three separate models: a prediction model, a baseline model and a selector model. The baseline model is trained to predict the true label $Y$ given the full feature vector $X$, and it can be trained independently of the remaining models; the predictor model makes predictions given subsets of features $X_S$ (with $S$ sampled according to the selector model), and it is trained to predict the true labels $Y$; and finally, the selector model takes a feature vector $X$ and outputs a probability distribution for a subset $S$.

The selector model, which ultimately outputs explanations, relies on the baseline model primarily for variance reduction purposes \citep{yoon2018invase}. Because the sampled subsets are used only for the predictor model, which is trained to predict the true label $Y$ (rather than the baseline model's predictions), we view the prediction model as the model being explained, and we understand it as removing features via a strategy of introducing missingness during training (Eq.~\ref{eq:missingness}).

For the optimization objective, \cite{yoon2018invase} explain that their aim is to minimize the following KL divergence for each input $x \in \mathcal{X}$:

\begin{equation*}
    S^* = \argmin_S \; \KL\big(p(Y \mid X = x) \; \big|\big| \; p(Y \mid X_S = x_S) \big) + \lambda |S|.
\end{equation*}

\noindent This is consistent with their learning algorithm if we assume that the predictor model outputs the Bayes optimal prediction $p(Y \mid X_S = x_S)$.
If we denote their predictor model as a subset function $F$ and interpret the KL divergence as a loss function with the true label $Y$ (i.e., cross entropy loss), then we can rewrite this objective as follows:

\begin{equation*}
    S^* = \argmin_S \; \E_{Y \mid X = x} \Big[ \ell\big(F(x_S), Y\big) \Big] + \lambda |S|.
\end{equation*}

\noindent This is the description of INVASE provided in the main text.



\subsection{REAL-X}

REAL-X \citep{jethani2021have} is similar to L2X and INVASE in that it uses a learned selector model to perform instance-wise feature selection. REAL-X is designed to resolve a flaw in L2X and INVASE, which is that both methods learn the selector model jointly with their subset functions, enabling label information to be leaked via the selected subset $S$. 

To avoid this issue, REAL-X learns a subset function $F$ independently from the selector model using the following objective function (with modified notation):

\begin{equation*}
    \min_F \; \E_{X} \E_{Y \sim f(X)} \E_{S} \Big[ \ell\big(F(X_S), Y\big) \Big].
\end{equation*}

\noindent The authors point out that $Y$ may be sampled from its true conditional distribution $p(Y \mid X)$ or from a model's distribution $Y \sim f(X)$; we remark that the former is analogous to INVASE (missingness introduced during training) and that the latter is analogous to L2X (training a surrogate model).
Notably, unlike L2X or INVASE, REAL-X optimizes its subset function with the subsets $S$ sampled independently from the input $X$, enabling it to approximate the Bayes optimal predictions $F(x_S) \approx p(Y \mid X_S = x_S)$.

We focus on the case with the label sampled according to $Y \sim f(X)$, which can be understood as fitting a surrogate model $F$ to the original model $f$. With the learned subset function $F$ fixed, REAL-X then learns a selector model that optimizes the following objective for each input $x \in \mathcal{X}$:

\begin{equation*}
    S^* = \argmin_S \; \E_{Y \sim f(x)} \Big[ \ell\big( F(x_S), Y \big) \Big] + \lambda |S|.
\end{equation*}

\noindent Rather than viewing this as the mean loss for labels sampled according to $f(x)$, we interpret this as a loss function between $F(x_S)$ and $f(x)$, as we did with L2X:

\begin{equation*}
    S^* = \argmin_S \; \ell\big( F(x_S), f(x) \big) + \lambda |S|.
\end{equation*}

\noindent This is our description of REAL-X provided in the main text.

\subsection{Prediction Difference Analysis (PredDiff)}

Prediction Difference Analysis \citep{zintgraf2017visualizing} removes individual features (or groups of features) and analyzes the difference in a model's prediction. Removed pixels are imputed by conditioning on their bordering pixels, which approximates sampling from the full conditional distribution $p(X_{\bar S} | X_S)$. Rather than measuring the prediction difference directly, the authors use attribution scores based on the log-odds ratio:

\begin{equation*}
    a_i = \log \frac{F(x)}{1 - F(x)} - \log \frac{F(x_{D \setminus \{i\}})}{1 - F(x_{D \setminus \{i\}})}.
\end{equation*}

\noindent We view this as another way of analyzing the difference in the model output for an individual prediction.

\subsection{Causal Explanations (CXPlain)}

CXPlain removes single features (or groups of features) for individual inputs and measures the change in the loss function \citep{schwab2019cxplain}. The authors propose calculating the attribution values

\begin{equation*}
    a_i(x) = \ell\big(F(x_{D \setminus \{i\}}), y\big) - \ell\big(F(x, y)\big)
\end{equation*}

\noindent and then computing the normalized values

\begin{equation*}
    w_i(x) = \frac{a_i(x)}{\sum_{j=1}^d a_j(x)}.
\end{equation*}

\noindent The normalization step enables the use of a learning objective based on Kullback-Leibler divergence for the explainer model, which is ultimately used to calculate attribution values in a single forward pass. The authors explain that this approach is based on a ``causal objective,''
but CXPlain is causal in the same sense as every other method described in our work.

\subsection{Randomized Input Sampling for Explanation (RISE)}

The RISE method \citep{petsiuk2018rise} begins by generating a large number of randomly sampled binary masks. In practice, the masks are sampled by dropping features from a low-resolution mask independently with probability $p$, upsampling to get an image-sized mask, and then applying a random jitter. Due to the upsampling, the masks have values $m \in [0, 1]^d$ rather than $m \in \{0, 1\}^d$.

The mask generation process induces a distribution over the masks, which we denote as $p(m)$. The method then uses the randomly generated masks to obtain a Monte Carlo estimate of the following attribution values:

\begin{equation*}
    a_i = \frac{1}{\E[M_i]} \E_{p(M)}\big[f(x \odot M) \cdot M_i\big].
\end{equation*}

\noindent If we ignore the upsampling step that creates continuous mask values, we see that these attribution values are the mean prediction when a given pixel is included:

\begin{align*}
    a_i &= \frac{1}{\E[M_i]} \E_{p(M)}\big[f(x \odot M) \cdot M_i\big] \\
    &= \sum_{m \in \{0, 1\}^d} f(x \odot m) \cdot m_i \cdot \frac{p(m)}{\E[M_i]} \\
    &= \E_{p(M | M_i = 1)}\big[f(x \odot M)\big].
\end{align*}


\subsection{Interactions Methods for Explanations (IME)}

IME was presented in two separate papers \citep{vstrumbelj2009explaining, vstrumbelj2010efficient}. In the original version, the authors recommended training a separate model for each subset of features. In the second version, the authors proposed the more efficient approach of marginalizing out the removed features from a single model $f$.

The latter paper is ambiguous about the specific distribution used to marginalize out held out features \citep{vstrumbelj2010efficient}. Lundberg and Lee~\cite{lundberg2017unified} view that features are marginalized out using their distribution from the training dataset (i.e., the marginal distribution). In contrast, Merrick and Taly~\cite{merrick2019explanation} view IME as marginalizing out features using a uniform distribution. Upon a close reading of the paper, we opt for the uniform interpretation, but the specific interpretation of IME's choice of distribution does not impact any of our conclusions.

\subsection{SHAP}

SHAP~\citep{lundberg2017unified} explains individual predictions by decomposing them with the game-theoretic Shapley value~\citep{shapley1953value}, similar to IME~\citep{vstrumbelj2009explaining, vstrumbelj2010efficient} and QII \citep{datta2016algorithmic}. The original work proposed marginalizing out removed features with their conditional distribution
but remarked that the joint marginal provided a practical approximation.
Marginalizing using the joint marginal distribution is now the default behavior. KernelSHAP is an approximation approach based on solving a weighted linear regression problem~\citep{lundberg2017unified}.

\subsection{TreeSHAP}

TreeSHAP uses a unique approach to handle held out features in tree-based models \citep{lundberg2020local}. It accounts for missing features using the distribution induced by the underlying trees, and, since it exhibits no dependence on the held out features, it is a valid extension of the original model. However, it cannot be viewed as marginalizing out features using a simple distribution. 

Given a subset of features, TreeSHAP makes a prediction separately for each tree and then combines each tree's prediction in the standard fashion. But when a split for an unknown feature is encountered, TreeSHAP averages predictions over the multiple paths in proportion to how often the dataset follows each path. This is similar but not identical to the conditional distribution because each time this averaging step is performed, TreeSHAP conditions only on coarse information about the features that preceded the split.

\subsection{LossSHAP}

LossSHAP is a version of SHAP that decomposes the model's loss for an individual prediction rather than the prediction itself. The approach was first considered in the context of TreeSHAP~\cite{lundberg2020local}, and it has been discussed in more detail as a local analogue to SAGE~\cite{covert2020understanding}.

\subsection{Shapley Net Effects}

Shapley Net Effects \citep{lipovetsky2001analysis} was originally proposed for linear models that use MSE loss, but we generalize the method to arbitrary model classes and arbitrary loss functions. Unfortunately, Shapley Net Effects quickly becomes impractical with large numbers of features or non-linear models.

\subsection{Shapley Effects}

Shapley Effects analyzes a variance-based measure of a function's sensitivity to its inputs, with the goal of discovering which features are responsible for the greatest variance reduction in the model output \citep{owen2014sobol}. The cooperative game described in the paper is:

\begin{equation*}
    u(S) = \Var\Big(\E\big[f(X) \; | \; X_S\big]\Big).
\end{equation*}

We present a generalized version to cast this method in our framework. In the appendix of Covert et al.~\cite{covert2020understanding}, it was shown that this game is equal to:

\begin{align*}
    u(S) &= \Var\Big(\E\big[f(X) \; | \; X_S\big]\Big) \\
    &= \Var\big(f(X)\big) - \E\Big[\Var\big(f(X) \; | \; X_S\big)\Big] \\
    &= c - \E\Big[\ell\big(\E\big[f(X) \; | \; X_S\big], f(X)\big)\Big] \\
    &= c - \underbrace{\E\Big[\ell\big(F(X_S), f(X)\big)\Big]}_{\mathclap{\text{Dataset loss w.r.t. output}}}.
\end{align*}

This derivation assumes that the loss function $\ell$ is MSE and that the subset function $F$ is $F(x_S) = \E[f(X) \; | \; X_S = x_S]$. Rather than the original formulation, we present a cooperative game that is equivalent up to a constant value and that provides flexibility in the choice of loss function:

\begin{equation*}
    w(S) =  - \E\Big[\ell\big(F(X_S), f(X)\big)\Big].
\end{equation*}

\clearpage
\bibliographystyle{plain}
\bibliography{neurips_main}

\end{document}